\documentclass[conference]{IEEEtran}

\usepackage[cmex10]{amsmath}
\usepackage{amsthm}
\usepackage{amssymb}
\usepackage{mathrsfs}
\usepackage{graphicx}
\usepackage{float}
\usepackage{array}
\usepackage{epstopdf}
\usepackage{multirow}
\usepackage{cite}

\begin{document}

\title{Deep-Sentiment: Sentiment Analysis Using
Ensemble of CNN and Bi-LSTM Models}

\author{Shervin Minaee$^*$, Elham Azimi$^*$, AmirAli Abdolrashidi$^{\dagger}$  \\
$^*$New York University
\\ $^{\dagger}$University of California, Riverside\\ \\
}

\maketitle

\begin{abstract}
With the popularity of social networks, and e-commerce websites, sentiment analysis has become a more active area of research in the past few years. 
On a high level, sentiment analysis tries to understand the public opinion about a specific product or topic, or trends from reviews or tweets. 
Sentiment analysis plays an important role in better understanding customer/user opinion, and also extracting social/political trends.
There has been a lot of previous works for sentiment analysis, some based on hand-engineering relevant textual features, and others based on different neural network architectures.
In this work, we present a model based on an ensemble of  long-short-term-memory (LSTM), and convolutional neural network (CNN), one to capture the temporal information of the data, and the other one to extract the local structure thereof. 
Through experimental results, we show that using this ensemble model we can outperform both individual models.
We are also able to achieve a very high accuracy rate compared to the previous works.
\end{abstract}

\IEEEpeerreviewmaketitle

\section{Introduction}
\label{sec:Intro}
Emotions exist in all forms of human communication. In many cases, they shape one's opinion of an experience, topic, event, etc. We can receive opinions and feedback for many products, online or otherwise, through various means, such as comments, reviews, and message forums, each of which can be in the form of text, video, polls and so on. 
One can find some type of sentiment in every type of feedback, e.g. if the overall experience is positive, negative, or neutral. The main challenge for a computer is to understand the underlying sentiment in all these opinions. 
Sentiment analysis involves the extraction of emotions from and classification of data, such as text, photo, etc., based on what they are conveying to the users \cite{ontosentic}. This can be used in dividing user reviews and public opinion on products, services, and even people, into positive/negative categories, detecting tones such as stress or sarcasm in a voice, and even finding bots and troll accounts on a social network \cite{darpa_bot}. 
While it is quite easy in some cases (e.g. if the text contains a few certain words), there are many factors to be considered in order to extract the overall sentiment, including those transcending mere words. 

In the age of information, the Internet, and social media, the need to collect and analyze sentiments has never been greater.
With the massive amount of data and topics online, a model can gather and track information about the public opinion regarding thousands of topics at any moment. This data can then be used for commercial, economical and even political purposes, which makes sentiment analysis an extremely important feedback mechanism.

Sentiment analysis and opinion annotation has started to grow more popular since the early 2000s \cite{sentiment_survey, earlywork1, earlywork2, earlywork3, earlywork4}. 
Here we mention some of the promising previous works. 
Cardie \cite{earlywork1} proposed an annotation scheme to support answering questions from different points of view and evaluated them. 
Pang \cite{earlywork3} proposed a method to gather the overall sentiment from movie reviews (positive/negative) using learning algorithms such as Bayesian and support vector machines (SVMs). 
Liu \cite{earlywork4} uses a real-world knowledge bank to map different scenarios to their corresponding emotions in order to extract an overall sentiment from a statement.

With the rising popularity of deep learning in the past few years, combined with the vast amount of labeled data, deep learning models have replaced many of the classical techniques used to solve various natural language processing and computer vision tasks. 
In these approaches, instead of extracting hand-crafted features from text and images, and feeding them to some classification model, end-to-end models are  used to jointly learn the feature representation and perform classification.
Deep learning based models have been able to achieve state-of-the-art performance on several tasks, such as sentiment analysis, question answering, machine translation, word embedding, and name entity recognition in NLP \cite{deep_nlp1}-\cite{deep_nlp7}, and image classification, object detection, image segmentation, image generation, and unsupervised feature learning in computer vision \cite{deep_vis1}-\cite{deep_vis7}.

Similar to other tasks mentioned above, there have been numerous works applying deep learning models to sentiment analysis in recent years \cite{deepwork1, deepwork2,deepwork3,deepsent1,deepsent2,deepsent3,deepsetnsurvey}. 
In \cite{deepwork1}, Santos et al proposed a sentiment analysis algorithm based on deep convolutional networks,  from short sentences and tweets. 
You et al \cite{deepwork2} used transfer learning approach for sentiment analysis, by progressively training them on some labeled data.
In \cite{deepwork3}, Lakkaraju et al proposed a hierarchical deep learning approach for aspect-specific sentiment analysis.
For a more comprehensive overview of deep learning based sentiment analysis, we refer the readers to \cite{deepsetnsurvey}.

In this paper, we seek to improve the accuracy of sentiment analysis using an ensemble of CNN and bidirectional LSTM (Bi-LSTM) networks, and test them on popular sentiment analysis databases such as the IMDB review and SST2 datasets. 
The block-diagram of the proposed algorithm is shown in Figure 1.
\begin{figure}[h]
\begin{center}
   \includegraphics[page=1,width=0.95\linewidth]{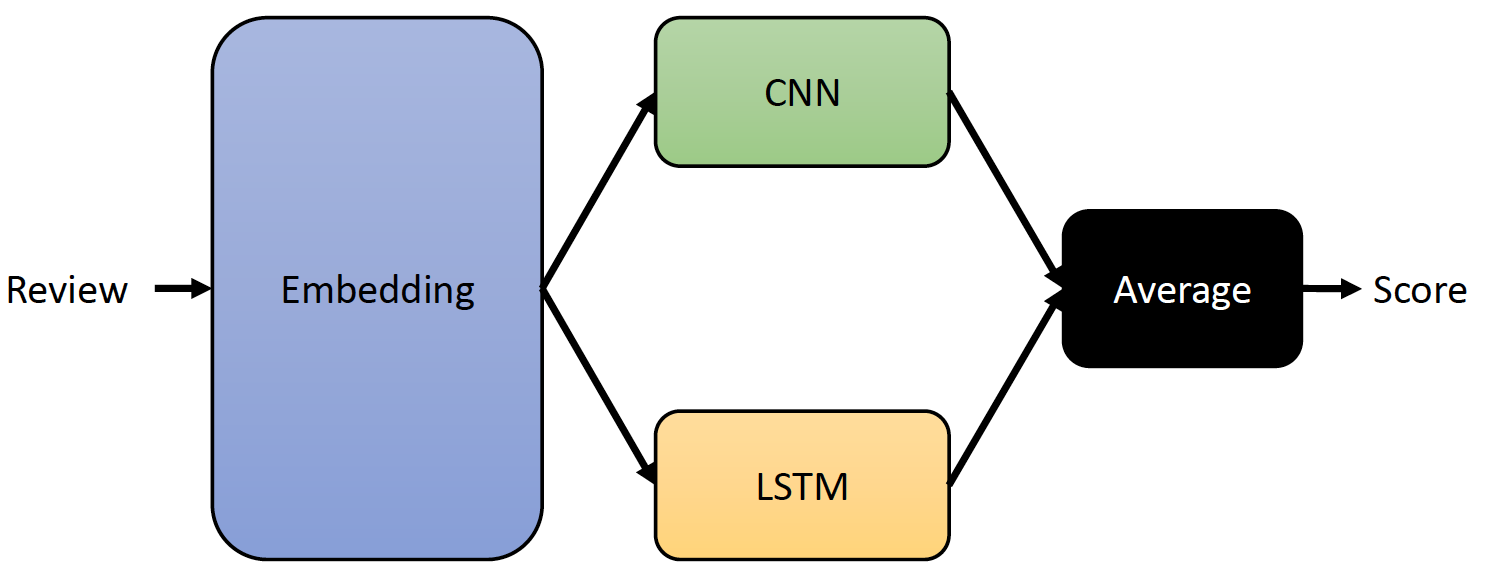}
\end{center}
   \caption{The block diagram of the proposed ensemble model.}
\label{fig:lstm_model}
\end{figure}

The CNN network tries to extract information about the local structure of the data by applying multiple filters (each having different dimensions), while the LSTM network is better suited to extract the temporal correlation of the data and dependencies in the text snippet. 
We then combined the predicted scores of these two models, to infer the sentiment of reviews.
Through experimental results, we show that by using an ensemble model, we are able to outperform the performance of both individual models (CNN, and LSTM).

The structure of the rest of this paper is as follows.
In Section II, we provide the detail of the proposed model, and the architecture for CNN and LSTM models used in our work. 
In Section III, we provide the results from our  experimental studies on two sentiment analysis databases. 
And finally we conclude the paper in Section IV.

\section{The Proposed Framework}
\label{sec:Framework}
As mentioned previously, we propose a framework based on the ensemble of LSTM and CNN models to perform sentiment analysis. Ensemble models have been used for various problems in NLP and vision, and are shown to bring performance gain over single models \cite{ens1}, \cite{ens2}.
In the following subsections we provide an overview of the proposed LSTM and CNN models.

\subsection{The LSTM model architecture}
LSTM \cite{lstm} is a popular recurrent neural network architecture for modeling sequential data, which is designed to have a better ability to capture long term dependencies than the vanilla RNN model.
As other kind of recurrent neural networks, at each time-step, LSTM network gets the input from the current time-step and the output from the previous time-step, and produces an output which is fed to the next time step.  
The hidden layer from the last time-step (and sometimes all hidden layers), are then used for classification.
The high-level architecture of a LSTM network is shown in Figure 2. 
\begin{figure}[h]
\begin{center}
   \includegraphics[page=1,width=0.8\linewidth]{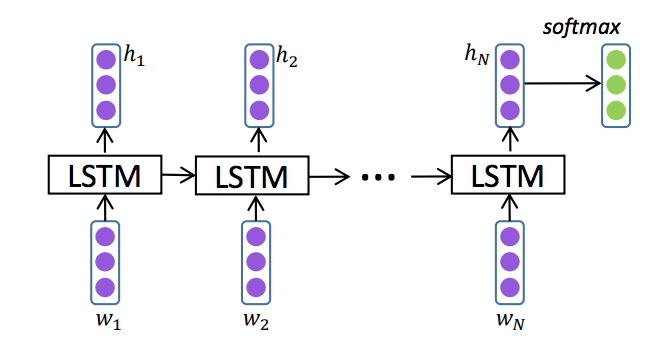}
\end{center}
   \caption{The architecture of a standard LSTM model \cite{lstm_cell}}
\label{fig:lstm_model}
\end{figure}

As mentioned above, the vanilla RNN often suffers from the gradient vanishing or exploding problems, and LSTM netowkr tries to overcome this issue by introducing some internal gates.
In the LSTM architecture, there are three gates (input gate, output gate, forget gate) and a memory cell.
The cell remembers values over arbitrary time intervals and the three gates regulate the flow of information into and out of the cell.
Figure 3 illustrates the inner architecture of a single LSTM module.
\begin{figure}[h]
\begin{center}
   \includegraphics[page=1,width=0.98\linewidth]{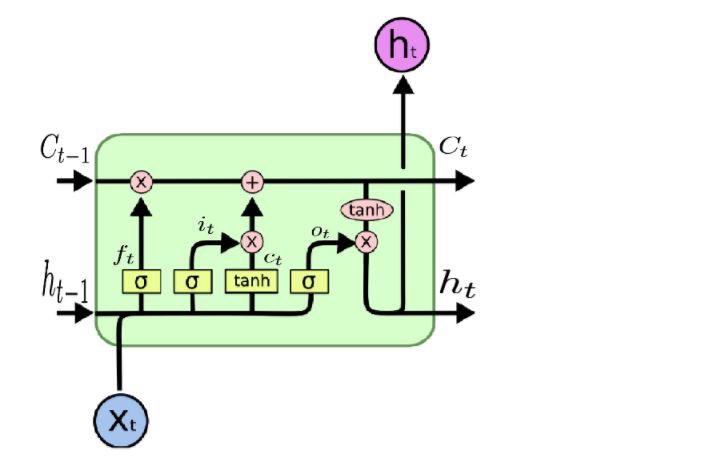}
\end{center}
   \caption{The architecture of a standard LSTM module \cite{lstm_cell}}
\label{fig:lstm_model}
\end{figure}

The relationship between input, hidden states, and different gates can be shown as:
\begin{equation}
\begin{aligned}
f_t= \sigma (\textbf{W}^{(f)} x_t+\textbf{U}^{(f)} h_{t-1}+ b^{(f)}  ), \hspace{2.73cm} \\
i_t= \sigma (\textbf{W}^{(i)} x_t+\textbf{U}^{(i)} h_{t-1}+ b^{(i)}  ), \hspace{2.94cm} \\
o_t= \sigma (\textbf{W}^{(o)} x_t+\textbf{U}^{(o)} h_{t-1}+ b^{(o)}  ), \hspace{2.78cm} \\
c_t= f_t \odot c_{t-1}+ i_t \odot \text{tanh} (\textbf{W}^{(c)} x_t+\textbf{U}^{(c)} h_{t-1}+ b^{(c)}  ), \\
h_t= o_t \odot \text{tanh}(c_t) \hspace{5.21cm}
\end{aligned}
\end{equation}
where $x_t \in R^d$ is the input at time-step t, and $d$ denotes the feature dimension for each word, $\sigma$ denotes the element-wise sigmoid function (to map the values within $[0,1]$),  $\odot$ denotes the element-wise product.
$c_t$ denotes the memory cell designed to lower the risk of vanishing/exploding gradient, and therefore enabling learning of dependencies over larger period of time feasible with traditional recurrent networks.
The forget gate, $f_t$ is to reset the memory cell. 
$i_t$ and $o_t$ denote the input and output gates, and essentially control the input and output of the memory cell.

For many applications, we are interested in the temporal information flow in both directions, and there is variant of LSTM, called Bidirectional-LSTM (Bi-LSTM), which can address this.
Bidirectional LSTMs train two hidden layers on the input sequence. The first one on the input sequence as-is, and the second one on the reversed copy of the input sequence. This can provide additional context to the network, by looking at both past and future information, and results in faster and better learning.

In our work, we used a two-layer bi-LSTM, which gets the Glove embedding \cite{glove} of words in a review, and predicts the sentiment for that. 
The architecture of the proposed Glove+bi-LSTM model is shown in Figure 4.
\begin{figure}[h]
\begin{center}
   \includegraphics[page=1,width=0.98\linewidth]{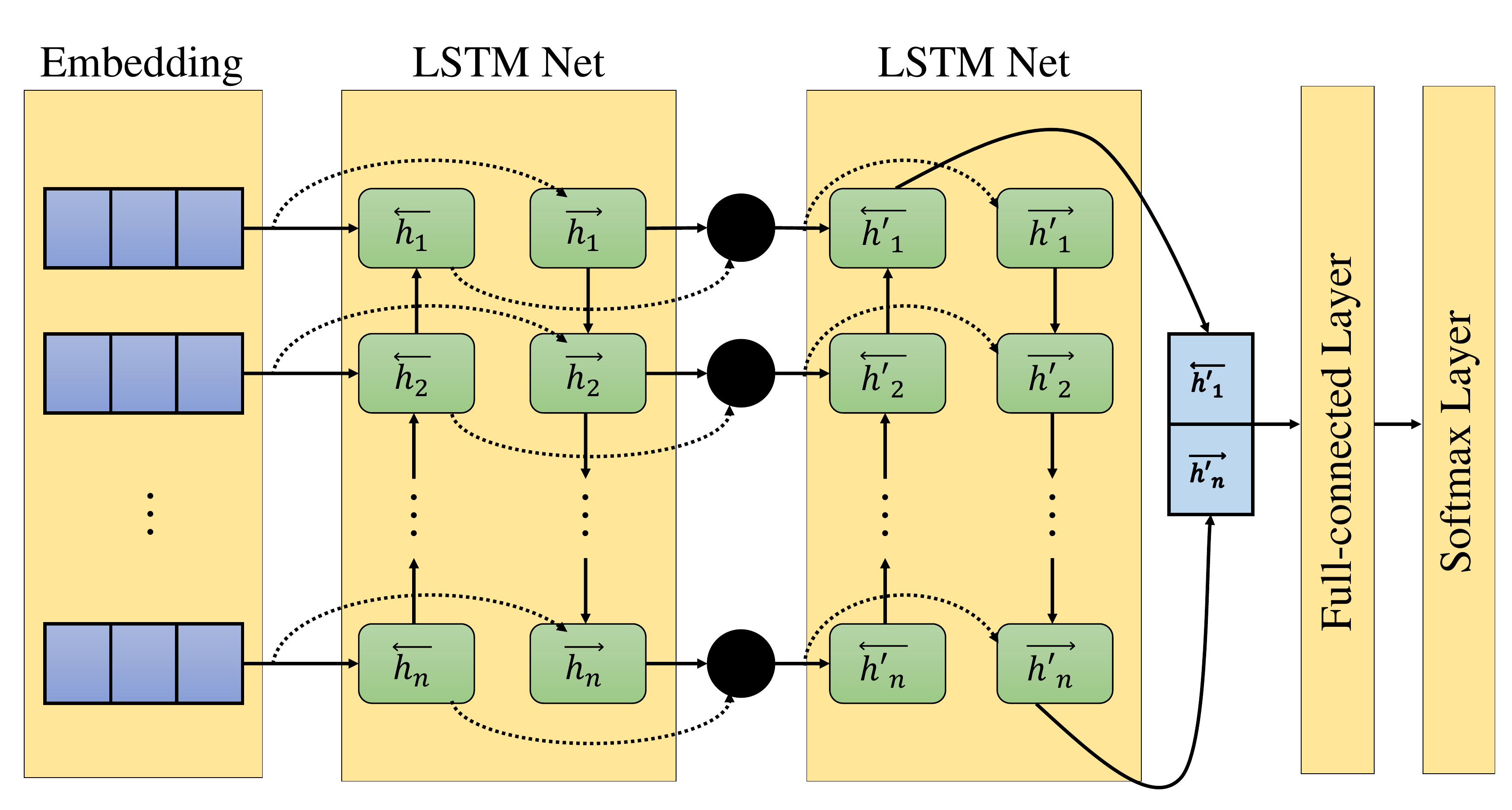}
\end{center}
   \caption{The architecture of a standard LSTM module \cite{lstm_cell}}
\label{fig:lstm_model}
\end{figure}

\subsection{The CNN model architecture}
Another piece of our proposed framework is based on convolutional neural network (CNN) \cite{cnn}. 
CNNs have been very successful for several computer vision and NLP tasks in the recent years.
They are specially powerful in exploiting the local correlation and pattern of the data through learned by their feature maps. 
One of the early works which used CNN for text classification is by Kim \cite{CNNtext}, which showed great performance on several text classification tasks.

To perform text classification with CNN, usually the embedding from different words of a sentence (or paragraph) are stacked together to form a two-dimensional array, and then convolution filters (of different length) are applied to a window of $h$ words to produce a new feature representation. 
Then some pooling (usually max-pooling) is applied on new features, and the pooled features from different filters are concatenated with each other to form  the hidden representation. 
These representations are then followed by one (or multiple) fully connected layer(s) to make the final prediction. 

In our work, we use word embedding based on the pre-trained Glove model, and used a convolutional neural network with 4 filter sizes (1,2,3,4), and 100 feature maps for each filter. 
The hidden representation are then followed by two fully-connected layers and fed into a softmax classifier.
The architecture of the proposed CNN model is shown below:
The general architecture of the proposed CNN model is shown in Figure 5.
\begin{figure}[h]
\begin{center}
   \includegraphics[page=2,width=0.98\linewidth]{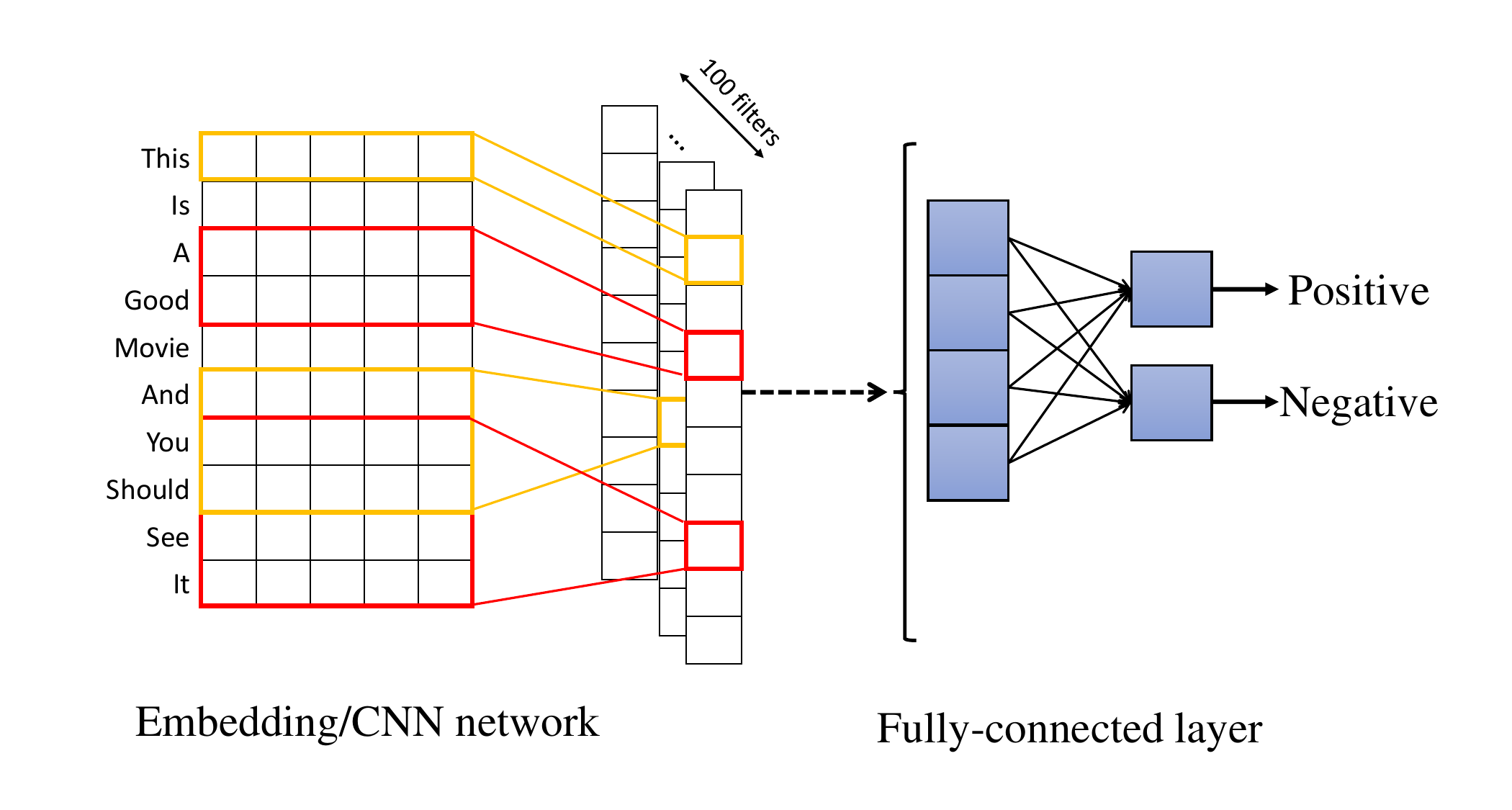}
\end{center}
   \caption{The general architecture CNN based text classification models}
\label{fig:cnn_model}
\end{figure}

\subsection{The Ensemble Model}
Both LSTM and CNN models perform reasonably well, and achieve good performance for sentiment analysis. LSTM achieves this mainly by looking at temporal information of data, and CNN by looking at the holistic view of local-information in text.
But we believe we can boost the performance further by combining the scores from these two models. 
In our work, we use an ensemble of CNN and LSTM models, by taking the average probability scores of these models as the final predictions. 
The overall architecture of our proposed model is shown in Figure 1.

\section{Experimental Results}
\label{sec:Evaluation}
Before presenting the experimental results, let us first discuss the hyper-parameters used in our training, and give an overview of the datasets used in our experiments.
We train the proposed CNN and LSTM models for 100 epochs using an Nvidia Tesla GPU.
The batch size is set to 64 for SST2 dataset, and to 50 for IMDB dataset. 
ADAM optimizer is used to optimize the loss function, with a learning rate of 0.0001 and a weight decay of 0.00001.
We present the details of the datasets used for our work in the following section.

\subsection{Databases}
In this section, we provide a quick overview of the datasets used in our work, IMDB review \cite{imdb}, and SST2 dataset \cite{sst2}.

\textbf{IMDB review dataset}:
IMDB dataset contains movie reviews along with their associated binary sentiment polarity labels. 
It contains 50,000 reviews split evenly into 25k train and 25k test sets. 
The overall distribution of labels is balanced (25k positive and 25k negative).
In the entire collection, no more than 30 reviews are allowed for any given movie to reduce correlation among reviews. Further, the train and test sets contain a disjoint set of movies.

\begin{figure}[h]
\begin{center}
   \includegraphics[width=0.95\linewidth]{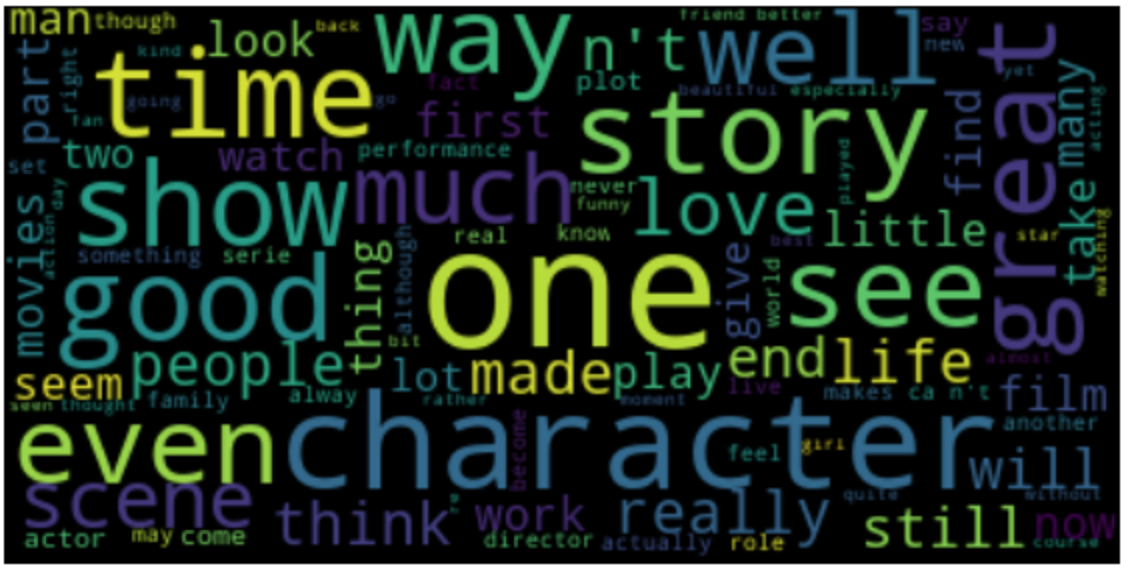} \\ \vspace{0.2cm}
   \includegraphics[width=0.95\linewidth]{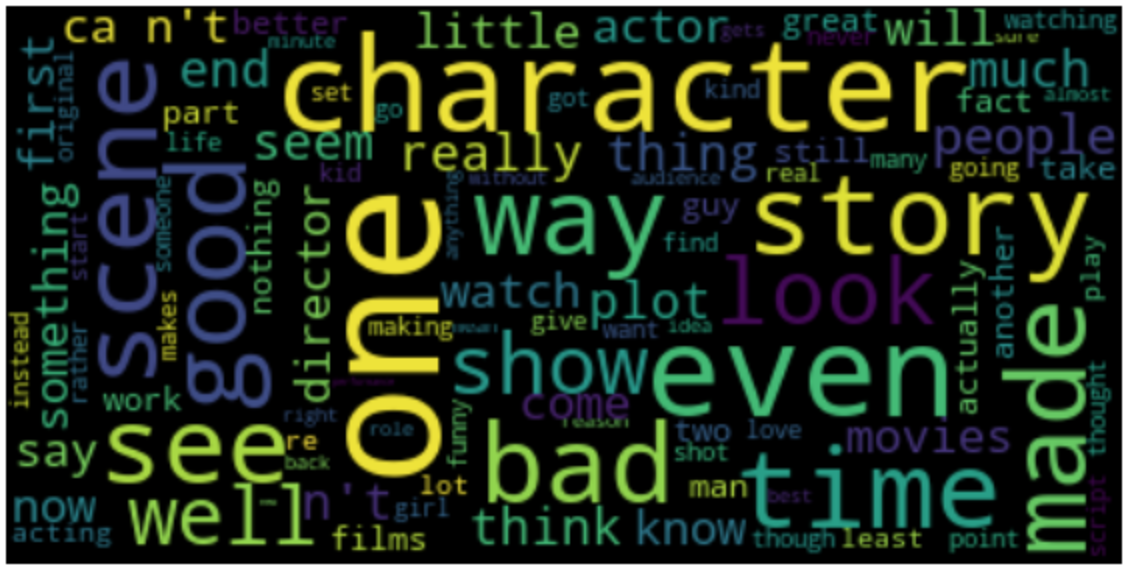}
\end{center}
   \caption{The visualization of most frequent words for both positive and negative reviews for IMDB database. The top and bottom image denote the frequent words for positive and negative reviews respectively.}
\label{fig:imdb}
\end{figure}

We illustrate some of the frequent words of this dataset in Figure \ref{fig:imdb}. 
It is worth mentioning that we excluded the stop words, as well as the words "movie" and "film" for this representation, as they do not carry a lot of information toward sentiment.
Some of the most distiguishable  words in the positive reviews include {"great", "well", "love"}, while words in the negative reviews include {"bad", "can't"}.
Some of the words are also common between both negative and positive classes, such as {"One", "Character"}, even the word "well" seems to be a common word in both negative and positive reviews, but it is more frequent in positive class (based on the size comparison of this word in two classes).

\textbf{SST2 Dataset}:
Stanford Sentiment Treebank2 (SST2) is a binary sentiment analysis dataset, with train/dev/test splits provided.
It is worth to mention that this data is actually provided at the phrase-level and hence one can train the model on both phrases and sentences. The vocabulary size for this dataset is around 16k.

\begin{figure}[h]
\begin{center}
   \includegraphics[width=0.95\linewidth]{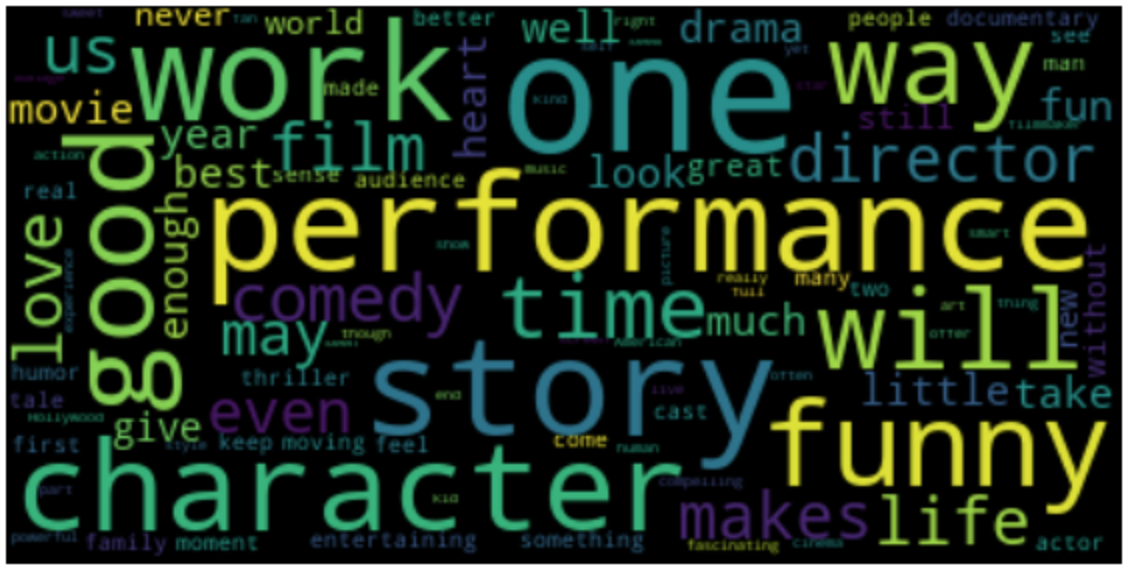} \\ \vspace{0.2cm}
   \includegraphics[width=0.95\linewidth]{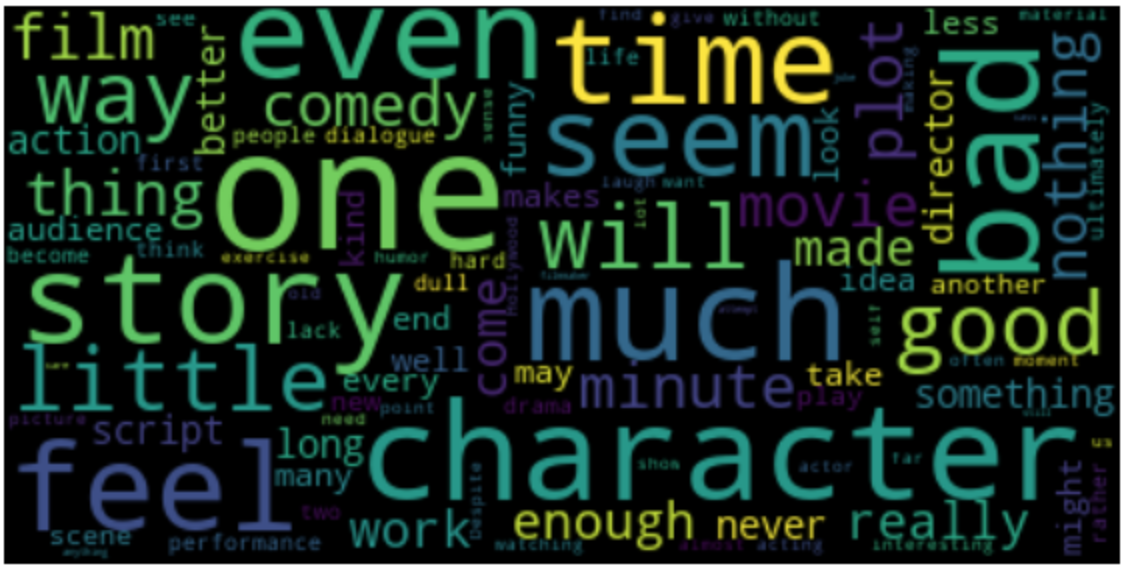}
\end{center}
   \caption{The visualization of most frequent words for both positive and negative reviews for SST2 database. The top and bottom image denote the frequent words for positive and negative reviews respectively.}
\label{fig:sst2}
\end{figure}

In Figure \ref{fig:sst2}, we show some of the most common words in SST2 dataset for both positive and negative reviews as a Wordcloud (we excluded the stop words, as well as the words "movie" and "film" for this representation, as mentioned above for IMDB dataset).
Some of the most distiguishable  words in the positive reviews include {"good", "funny", "love", "best"}, while words in the negative reviews include {"bad","nothing","never"}.
Some of the words are also common between both negative and positive classes, such as {"one", "character", "way"}, as well "movie" which is removed from the set of words.

\subsection{Model Performance and Comparison}
We will now present the experimental results of the proposed ensemble model on the above datasets.

We first compare the classification accuracy of the CNN and LSTM models, with the ensemble of these two.
The classification accuracies for the CNN+Glove, LSTM+Glove, as well as the ensemble of these two models on IMDB, and SST2 dataset are presented in Table \ref{table:imdb} and Table \ref{table:SST2}  respectively.
\begin{table}[h]
\centering
  \caption{Model Performance on IMDB Dataset}
\begin{tabular}{|m{4cm}|m{2cm}|}
\hline
Method  & Accuracy Rate\\
\hline 
Proposed LSTM Model &   \ \ \  89\% \\
\hline 
Proposed CNN Model &   \ \ \  89.3\% \\
\hline 
Proposed Ensemble of LSTM and CNN &  \ \ \ 90\% \\
\hline
\end{tabular}
\label{table:imdb}
\end{table}

\begin{table}[h]
\centering
  \caption{Model Performance on SST2 Dataset}
\begin{tabular}{|m{4cm}|m{2cm}|}
\hline
Method  & Accuracy Rate\\
\hline 
Proposed LSTM Model &   \ \ \  80\% \\
\hline 
Proposed CNN Model &   \ \ \  80.2\% \\
\hline 
Proposed Ensemble of LSTM and CNN &  \ \ \ 80.5\% \\
\hline
\end{tabular}
\label{table:SST2}
\end{table}
As we can see, we achieve some performance gain by using the ensemble model over the individual ones. 

In Table \ref{table:imdb_comp} we  provide the comparison between the proposed algorithm, and some of the previous works on IMDB review sentiment analysis. 
\begin{table}[h]
\centering
  \caption{The performance comparison with some of the previous works on IMDB dataset}
\begin{tabular}{|m{5cm}|m{1.5cm}|}
\hline
Method  & Accuracy Rate\\
\hline
LDA  \cite{WRRBM} &   \ \ \  67.42\% \\
\hline 
LSA  \cite{WRRBM} &   \ \ \  83.96\% \\
\hline 
Semantic + Bag of Words (bnc  \cite{WRRBM} &   \ \ \  88.28\% \\
\hline 
SA-LSTM with joint training  \cite{semi_supervised} &   \ \ \  85.3\% \\
\hline
LSTM with tuning and dropout  \cite{semi_supervised} &   \ \ \  86.5\% \\
\hline
SVM on BOW \cite{bow} &   \ \ \  88.7\% \\
\hline
WRRBM + BoW (bnc)  \cite{WRRBM} &   \ \ \  89.3\% \\
\hline 
Proposed Ensemble of LSTM and CNN &  \ \ \ 90\% \\
\hline
\end{tabular}
\label{table:imdb_comp}
\end{table}


\subsection{The predicted scores for positive and negative reviews}
\begin{figure}[h]
\begin{center}
   \includegraphics[width=0.9\linewidth]{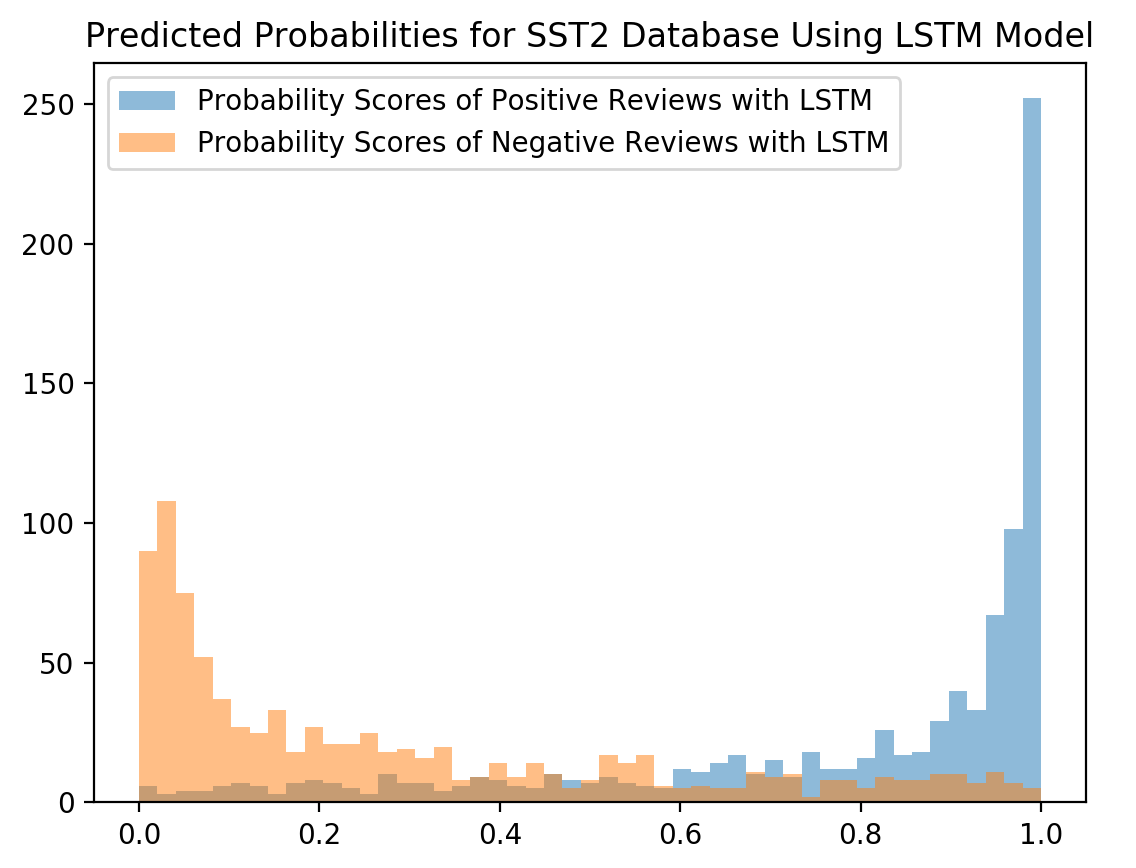}
   \includegraphics[width=0.9\linewidth]{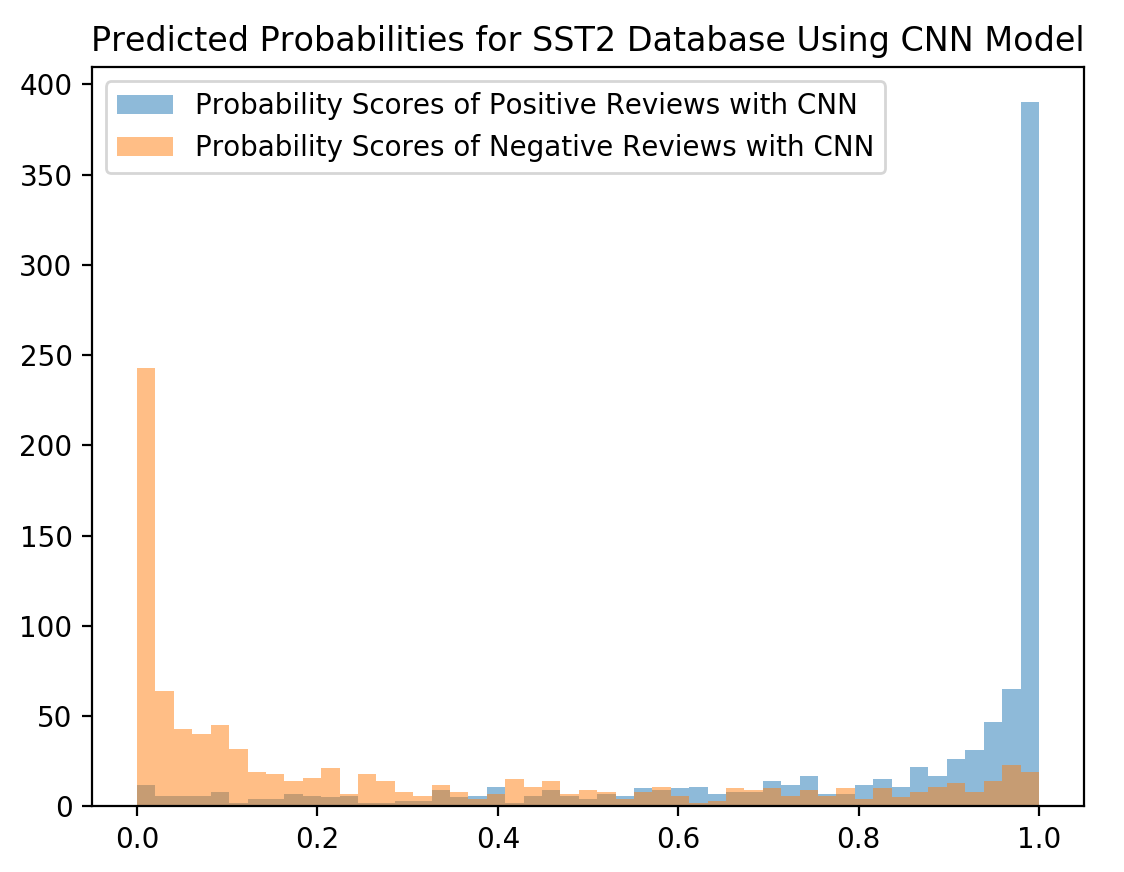}
\end{center}
   \caption{The distribution of predicted probability scores predicted by LSTM and CNN for the reviews in SST2 database.}
\label{fig:SST2_score_hist}
\end{figure}

In this section we present the distribution of probability scores predicted by LSTM and CNN models for the reviews in SST2 database.
Ideally, we expect the predicted scores for positive and negative reviews to be close to 1 and 0 respectively. 
The histogram of the predicted scores by our CNN and LSTM models (for SST2 dataset) are presented in Figure \ref{fig:SST2_score_hist}. 

As we can CNN model performs slightly better in pushing the scores for positive and negative reviews toward 1 and 0, which results in slightly higher accuracy. For LSTM model, the predicted probabilities cover a wide range of values in $[0,1]$, which makes it challenging to find a good cut-off threshold to separate two classes.
It is worth mentioning that in our current models, 0.5 is used for the cut-off thresholds between positive and negative classes, but perhaps this can be further improved by tuning this threshold on a validation set.

\subsection{Training Performance over Time}
\begin{figure}[h]
\begin{center}
   \includegraphics[width=0.9\linewidth]{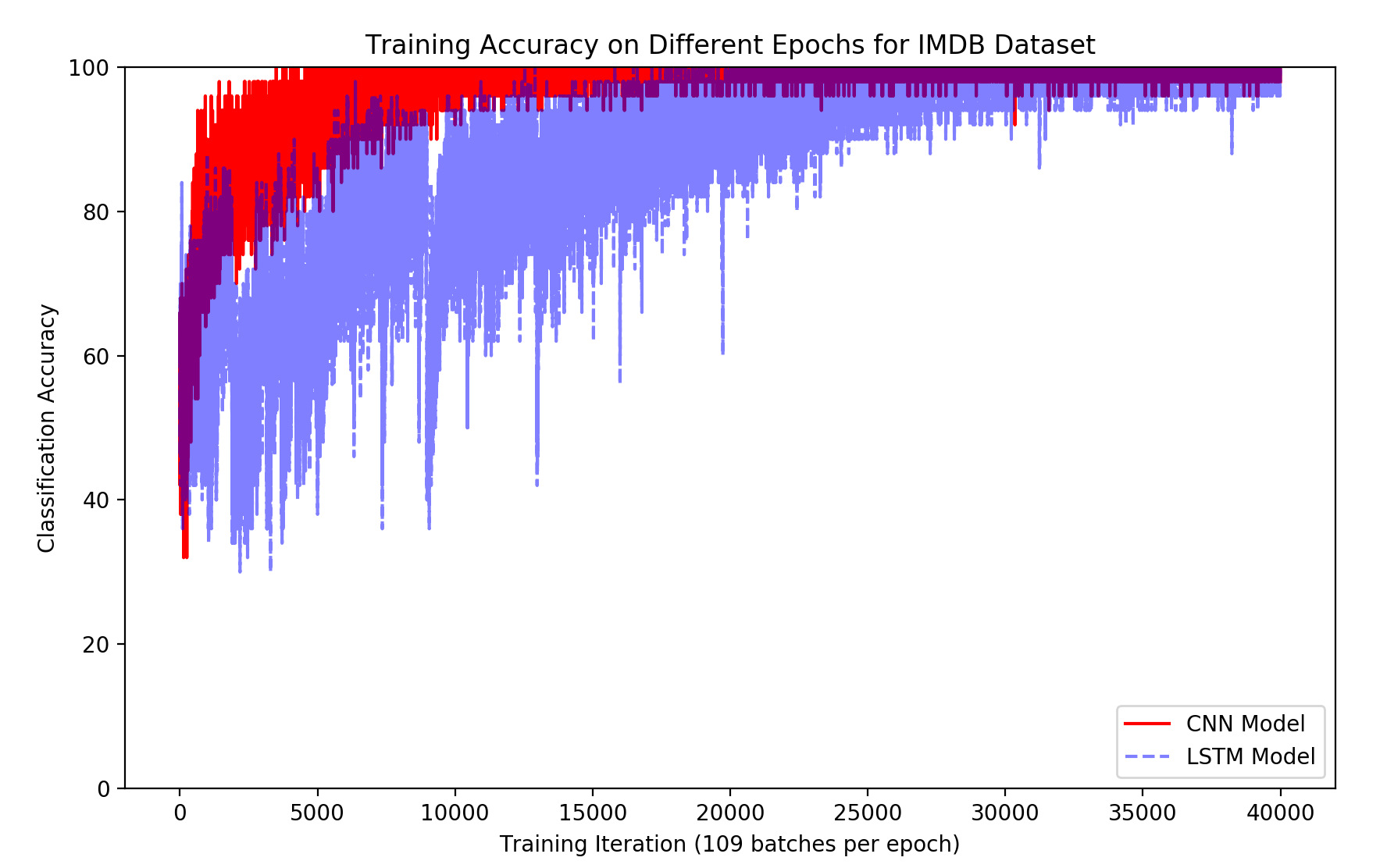}
\end{center}
   \caption{The training accuracy at different epochs for IMDB database}
\label{fig:imdb_train}
\end{figure}

\begin{figure}[h]
\begin{center}
   \includegraphics[width=0.9\linewidth]{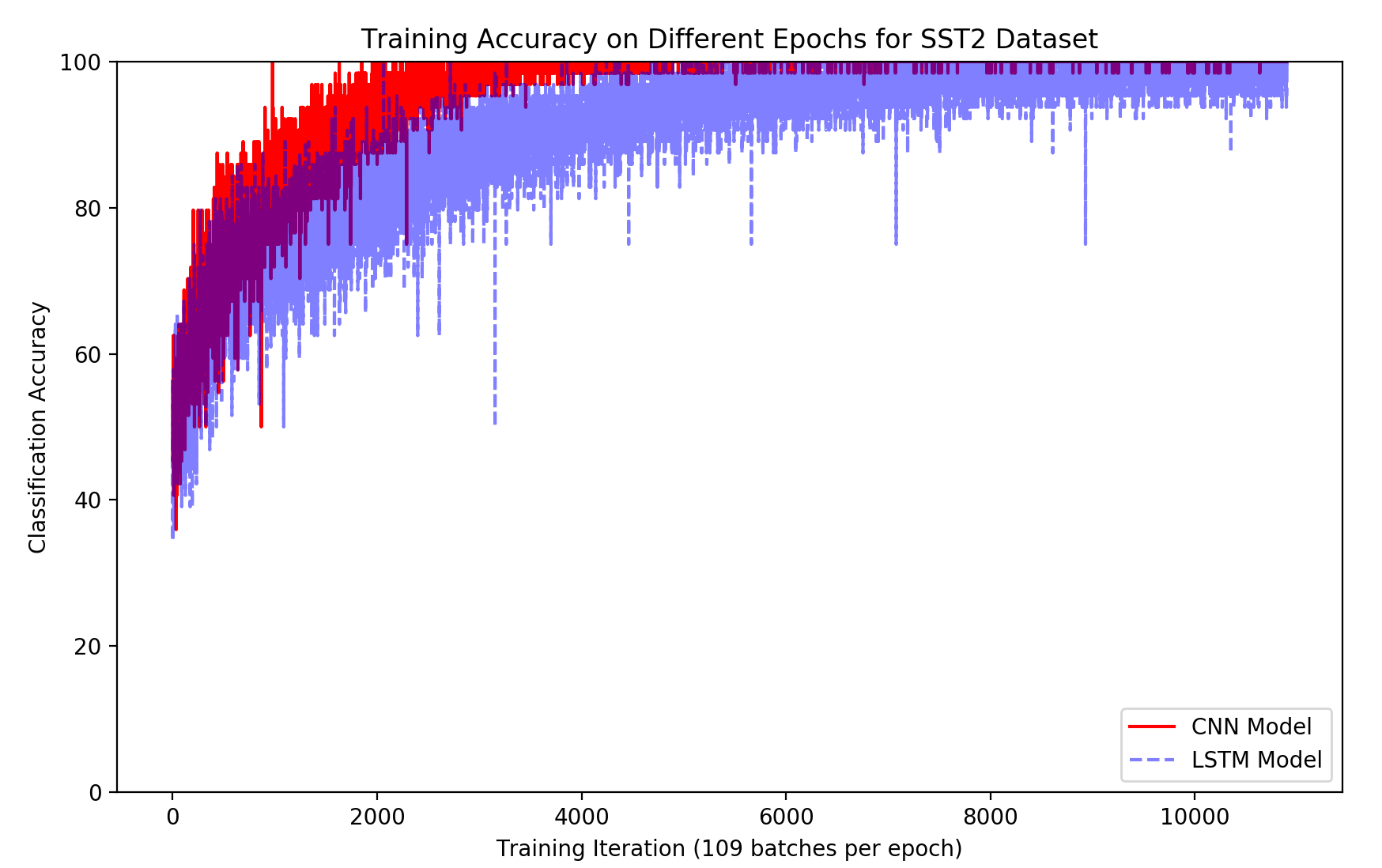}
\end{center}
   \caption{The training accuracy at different epochs for SST2 database}
\label{fig:sst2_train}
\end{figure}

We also present the training accuracies for both CNN and LSTM models, on IMDB and SST2 datasets over different epochs. These accuracies are shown in Figures \ref{fig:imdb_train} and \ref{fig:sst2_train} respectively.
As we can see the CNN model starts converging earlier than LSTM on both datasets.

\section{Conclusion}
\label{sec:Conclusion}
In this work we proposed a framework for sentiment analysis, based on an ensemble of LSTM and CNN models.
Each word in reviews are represented with Glove embedding, and the embedding are then fed into the CNN and LSTM  models for prediction.
The predicted scores by LSTM and CNN model are then averaged to make the final predictions. 
Through experimental studies, we observe some performance gain by the ensemble model compared to the individual LSTM and CNN models. 
As a future work, we plan to jointly train the LSTM and CNN model, to further improve their performance gain over individual models. 
We believe this ensemble model can be used to improve the prediction accuracy in several other deep learning-based text processing tasks.

\section*{Acknowledgment}
We would like to thank Stanford University for providing the SST2 dataset, and also IMDB Inc for making IMDB dataset available.


\begin{thebibliography}{1}
\small
\bibitem{ontosentic}
Dragoni, Mauro, Soujanya Poria, and Erik Cambria. "OntoSenticNet: A commonsense ontology for sentiment analysis." IEEE Intelligent Systems 33.3 (2018): 77-85.
\bibitem{darpa_bot}
Subrahmanian, V. S., et al. "The DARPA Twitter bot challenge." Computer 49.6 (2016): 38-46.

\bibitem{sentiment_survey}
Hu, Minqing, and Bing Liu. "Mining opinion features in customer reviews." AAAI, Vol. 4, No. 4, 2004.
\bibitem{earlywork1}
Cardie, Claire, et al. "Combining Low-Level and Summary Representations of Opinions for Multi-Perspective Question Answering." New directions in question answering. 2003.
\bibitem{earlywork2}
Das, Sanjiv, and Mike Chen. "Yahoo! for Amazon: Extracting market sentiment from stock message boards." Proceedings of the Asia Pacific finance association annual conference (APFA). Vol. 35. 2001.
\bibitem{earlywork3}
Pang, Bo, Lillian Lee, and Shivakumar Vaithyanathan. "Thumbs up?: sentiment classification using machine learning techniques." Proceedings of the ACL-02 conference on Empirical methods in natural language processing-Volume 10. Association for Computational Linguistics, 2002.
\bibitem{earlywork4}
Liu, Hugo, Henry Lieberman, and Ted Selker. "A model of textual affect sensing using real-world knowledge." Proceedings of the 8th international conference on Intelligent user interfaces. ACM, 2003.

\bibitem{deep_nlp1}
Mikolov, Tomas, Ilya Sutskever, Kai Chen, Greg S. Corrado, and Jeff Dean. "Distributed representations of words and phrases and their compositionality." In Advances in neural information processing systems, pp. 3111-3119, 2013.
\bibitem{deep_nlp2}
Sutskever, Ilya, Oriol Vinyals, and Quoc V. Le. "Sequence to sequence learning with neural networks." Advances in neural information processing systems, 2014.
\bibitem{deep_nlp3}
Bahdanau, Dzmitry, Kyunghyun Cho, and Yoshua Bengio. "Neural machine translation by jointly learning to align and translate", ICLR, 2015.
\bibitem{deep_nlp4}
Li, Jiwei, Will Monroe, Alan Ritter, Michel Galley, Jianfeng Gao, and Dan Jurafsky. "Deep reinforcement learning for dialogue generation", EMNLP, 2016.
\bibitem{deep_nlp5}
Santos, Cicero Nogueira dos, and Victor Guimaraes. "Boosting named entity recognition with neural character embeddings", Proceedings of NEWS 2015 The Fifth Named Entities Workshop, 2015.
\bibitem{deep_nlp6}
Minaee, Shervin, and Zhu Liu. "Automatic question-answering using a deep similarity neural network." Global Conference on Signal and Information Processing (GlobalSIP), IEEE, 2017.
\bibitem{deep_nlp7}
AM Rush, S Chopra, J Weston, "A neural attention model for abstractive sentence summarization." arXiv preprint arXiv:1509.00685, 2015..
\bibitem{deep_vis1}
Krizhevsky, Alex, Ilya Sutskever, and Geoffrey E. Hinton. "Imagenet classification with deep convolutional neural networks", Advances in neural information processing systems,  2012.
\bibitem{deep_vis2}
Long, Jonathan, Evan Shelhamer, and Trevor Darrell. "Fully convolutional networks for semantic segmentation." Proceedings of the IEEE conference on computer vision and pattern recognition. 2015.
\bibitem{deep_vis3}
W Liu, D Anguelov, D Erhan, C Szegedy, S Reed, CY Fu, and Berg, A. C. "SSD: Single shot multibox detector", In European conference on computer vision (pp. 21-37). Springer, 2016.
\bibitem{deep_vis4}
He, Kaiming, Georgia Gkioxari, Piotr Dollár, and Ross Girshick, "Mask R-CNN." In Proceedings of the IEEE international conference on computer vision, pp. 2961-2969. 2017.
\bibitem{deep_vis5}
Goodfellow, Ian, Jean Pouget-Abadie, Mehdi Mirza, Bing Xu, David Warde-Farley, Sherjil Ozair, Aaron Courville, and Yoshua Bengio, "Generative adversarial nets", In Advances in neural information processing systems (pp. 2672-2680), 2014.
\bibitem{deep_vis6}
Minaee, Shervin, Amirali Abdolrashidiy, and Yao Wang. "An experimental study of deep convolutional features for iris recognition." signal processing in medicine and biology symposium (SPMB), IEEE, 2016.
\bibitem{deep_vis7}
Minaee, S., Wang, Y., Aygar, A., Chung, S., Wang, X., Lui, Y.W., Fieremans, E., Flanagan, S. and Rath, J., "MTBI Identification From Diffusion MR Images Using Bag of Adversarial Visual Features" IEEE transactions on medical imaging, 2019.

\bibitem{deepwork1}
Dos Santos, Cicero, and Maira Gatti. "Deep convolutional neural networks for sentiment analysis of short texts." Proceedings of COLING 2014, the 25th International Conference on Computational Linguistics: Technical Papers. 2014.
\bibitem{deepwork2}
You, Quanzeng, et al. "Robust image sentiment analysis using progressively trained and domain transferred deep networks." Twenty-Ninth AAAI Conference on Artificial Intelligence. 2015.
\bibitem{deepwork3}
Lakkaraju, Himabindu, Richard Socher, and Chris Manning. "Aspect specific sentiment analysis using hierarchical deep learning." NIPS Workshop on deep learning and representation learning, 2014.
\bibitem{deepsent1}
Severyn, Aliaksei, and Alessandro Moschitti. "Twitter sentiment analysis with deep convolutional neural networks." Proceedings of the 38th International ACM SIGIR Conference on Research and Development in Information Retrieval. ACM, 2015.
\bibitem{deepsent2}
Araque, Oscar, et al. "Enhancing deep learning sentiment analysis with ensemble techniques in social applications." Expert Systems with Applications 77 (2017): 236-246.
\bibitem{deepsent3}
Badjatiya, Pinkesh, et al. "Deep learning for hate speech detection in tweets." Proceedings of the 26th International Conference on World Wide Web Companion. International World Wide Web Conferences Steering Committee, 2017.
\bibitem{deepsetnsurvey}
Zhang, Lei, Shuai Wang, and Bing Liu. "Deep learning for sentiment analysis: A survey." Wiley Interdisciplinary Reviews: Data Mining and Knowledge Discovery 8.4, 2018.

\bibitem{ens1}
Kanakaraj, Monisha, and Ram Mohana Reddy Guddeti. "Performance analysis of Ensemble methods on Twitter sentiment analysis using NLP techniques", International Conference on Semantic Computing, IEEE, 2015.
\bibitem{ens2}
Bosch, Anna, Andrew Zisserman, and Xavier Munoz. "Image classification using random forests and ferns." international conference on computer vision, IEEE, 2007.

\bibitem{lstm}
Hochreiter, S., Schmidhuber, J. "Long short-term memory", Neural computation, 9(8), 1735-1780, 1997.
\bibitem{lstm_cell}
http://colah.github.io/posts/2015-08-Understanding-LSTMs/
\bibitem{glove}
Pennington, Jeffrey, Richard Socher, and Christopher Manning. "Glove: Global vectors for word representation", conference on empirical methods in natural language processing (EMNLP). 2014.

\bibitem{cnn}
Y LeCun, L Bottou, Y Bengio, P Haffner, "Gradient-based learning applied to document recognition" Proceedings of the IEEE, 86(11), 2278-2324, 1998.
\bibitem{CNNtext}
Kim, Yoon. "Convolutional neural networks for sentence classification", EMNLP, 2014.


\bibitem{imdb}
https://www.kaggle.com/iarunava/imdb-movie-reviews-dataset
\bibitem{sst2}
https://nlp.stanford.edu/sentiment/

\bibitem{WRRBM}
AL Maas, RE Daly, PT Pham, D Huang, AY Ng, C. Potts. "Learning word vectors for sentiment analysis", In ACL, 2011.
\bibitem{semi_supervised}
A Dai, V. Le Quoc, "Semi-supervised sequence learning." Advances in neural information processing systems, pp. 3079-3087, 2015.
\bibitem{bow}
Johnson, Rie, and Tong Zhang. "Supervised and semi-supervised text categorization using LSTM for region embeddings." arXiv preprint arXiv:1602.02373, 2016.


\end{thebibliography}
\end{document}